# Machine learning in the prediction of cardiac epicardial and mediastinal fat volumes

É.O. Rodrigues [a, *], V.H.A. Pinheiro [b], P. Liatsis [c], A. Conci [a]

[a] *Department of Computer Science, Universidade Federal Fluminense, Niterói, Rio de Janeiro, Brazil*
[b] *Department of Computer Science, Universidade Federal de Santa Catarina, Florianópolis, Santa Catarina, Brazil*
[c] *Department of Electrical Engineering, The Petroleum Institute, PO Box 2533 Abu Dhabi, United Arab Emirates*

A R T I C L E   I N F O

A B S T R A C T

We propose a methodology to predict the cardiac epicardial and mediastinal fat volumes in computed tomography images using regression algorithms. The obtained results indicate that it is feasible to predict these fats with a high degree of correlation, thus alleviating the requirement for manual or automatic segmentation of both fat volumes. Instead, segmenting just one of them suffices, while the volume of the other may be predicted fairly precisely. The correlation coefficient obtained by the Rotation Forest algorithm using MLP Regressor for predicting the mediastinal fat based on the epicardial fat was 0.9876, with a relative absolute error of 14.4% and a root relative squared error of 15.7%. The best correlation coefficient obtained in the prediction of the epicardial fat based on the mediastinal was 0.9683 with a relative absolute error of 19.6% and a relative squared error of 24.9%. Moreover, we analysed the feasibility of using linear regressors, which provide an intuitive interpretation of the underlying approximations. In this case, the obtained correlation coefficient was 0.9534 for predicting the mediastinal fat based on the epicardial, with a relative absolute error of 31.6% and a root relative squared error of 30.1%. On the prediction of the epicardial fat based on the mediastinal fat, the correlation coefficient was 0.8531, with a relative absolute error of 50.43% and a root relative squared error of 52.06%. In summary, it is possible to speed up general medical analyses and some segmentation and quantification methods that are currently employed in the state-of-the-art by using this prediction approach, which consequently reduces costs and therefore enables preventive treatments that may lead to a reduction of health problems.

## 1. Introduction

Medical diagnosis support systems speed up the tedious and meticulous analysis done by physicians or technicians on medical data. In many cases, a huge amount of data has to be analysed and, therefore, the diagnosis or the support data may lack precision and suffer noticeable inter and/or intra-observer variation.

Cardiac epicardial and mediastinal fats are correlated to several cardiovascular risk factors [1]. At present, three techniques (i.e., modalities) appear suitable for the quantification of these adipose tissues, namely Magnetic Resonance Imaging (MRI), Echocardiography and Computed Tomography (CT). Computed tomography provides a more accurate evaluation of fat tissues due to its higher spatial resolution when compared to ultrasound and MRI. In addition, CT is also widely used for computing the coronary calcium score.

In this work, we propose and analyse different methods for predicting the amount of epicardial fat based on the mediastinal fat and vice-versa. This is important mainly to speed up the process of their segmentation, regardless of being a manual or automatic segmentation. In our previous work [1], these two fats were automatically segmented in approximately 1.8 h (in volumes of 50 images on average). If the segmentation of one of these fats is known *a priori*, the volume of the other fat can be predicted in real time. This implies that a patient scan can be processed in approximately half of the computing time.

This work is organized as follows: Section 2 presents an overview of the state-of-the-art, further reinforcing the clinical significance of estimating these two fat volumes. In Section 3, the methodology and data used in this work are introduced and discussed. Section 4 contains the performed experiments, comparisons and overall analyses. Finally, in the last section, the conclusions of this work are summarized and avenues for future research are suggested.

* Corresponding author.
*E-mail addresses:* erickr@id.uff.br (É.O. Rodrigues), vhugo92@gmail.com (V.H.A. Pinheiro), pliatsis@pi.ac.ae (P. Liatsis), aconci@ic.uff.br (A. Conci).

## 2. Literature review

The human heart (Fig. 1) is enclosed in the pericardium, a fibroserous sac comprising three concentric layers [1]. The outermost layer is a densely fibrous, tough and inelastic structure (fibrous pericardium). Inside the fibrous pericardium is the serous pericardium, which consists of two layers. The outer of these layers (which is firmly applied to the inner surface of the fibrous pericardium) is termed parietal layer. This layer is reflected around the roots of the major vessels to become continuous with the visceral layer (epicardium), which covers the internal surface of the heart and is firmly applied to it [2].

Two distinct types of fat are closely associated to the human heart, namely the epicardial and mediastinal fats. They are separated by the epicardium layer. In a sense, the epicardial fat is the fat located within the cardiac sac (pericardium), while the mediastinal is the fat that lays on the outer surface of the heart, within the mediastinum. Some authors refer to it as pericardial fat, and indeed there exists a slight divergence in the nomenclature on this matter, as previous discussed in [1].

Some studies [3,4] associate the amount of epicardial adipose tissue with the progression of coronary artery calcification. Schlett et al. [5], for instance, found that the epicardial fat volume is nearly twice as high in patients with high-risk coronary lesions as compared to those without coronary artery calcification. Several studies also correlate other cardiovascular risk factors and outcomes to the epicardial adipose tissue volume, such as diastolic filling [6], myocardial infarction [3], atrial fibrillation and ablation outcome [5], carotid stiffness [7], atherosclerosis [8–10], and many others [11,3,12–14]. Furthermore, Chen et al. [15] have also shown that a high coronary artery calcium score may be associated to a higher general cancer incidence.

In addition, some studies address the importance of the mediastinal fat and its correlation with pathogenic profiles, health risk factors and diseases [16,17]. Some works [10,7] associate mediastinal fat, along with epicardial fat, to carotid stiffness. Others [10,4] relate these fats to atherosclerosis and coronary artery calcification. Sicari et al. [18] have also shown how mediastinal fat, rather than epicardial fat, is a cardiometabolic risk marker.

Moreover, a 16-year study [19] that assessed a total of 384597 patients found a rate of approximately 38.4% of deaths in the subsequent 28 days for individuals that had their first major coronary event. The same study also concluded that the probability of fatalities is slightly smaller in women. Another study ranked cardiovascular incidents as the most common cause of sudden natural death [20]. Therefore, the practice of automatically evaluating the amount of fat related to the heart may significantly contribute in avoiding similar outcomes.

Automated quantitative analysis of the epicardial and mediastinal fats could also add a prognostic value to cardiac CT trials, thus improving its cost-effectiveness. Besides that, automation diminishes the variability introduced by different human observers. In fact, quantifying the amount of fats by direct user intervention is highly prone to inter- and intra-observer variability. Thus, the evaluated samples may not be segmented using a set of objective principles. Iacobellis et al. [21] have shown that epicardial fat thickness and coronary artery disease, for instance, correlate independently of obesity. This evidence supports the need for individual segmentation and quantification of the adipose tissues rather than merely estimating epicardial fat based on the overall fat of the patient.

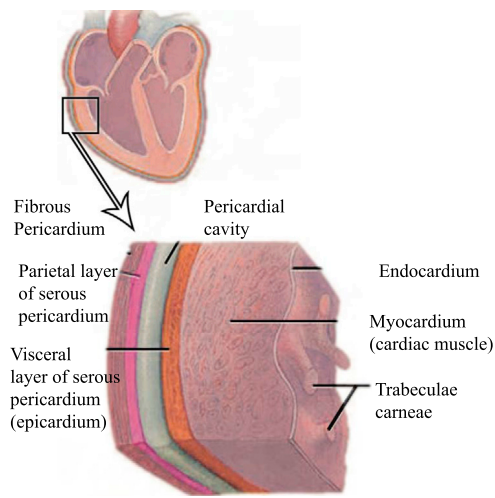

**Fig. 1.** A cross-section of the heart (top) and the associated cardiac layers (bottom).

In a previous work [1], we proposed an accurate method for automatically segmenting epicardial and mediastinal fats in CT exams. The method is mainly based on feature extraction [22] and classification algorithms [23]. However, registration is also performed for image alignment prior to feature extraction. Despite the superior accuracy of the results, when compared to the state-of-the-art, the downside of the approach is its processing time. Currently, it is able to automatically segment the cardiac fats (epicardial and mediastinal) of a patient (a set of approximately 50 images) in approximately 1.8 h.

In this work, we employ regression algorithms to predict the amount of one of the cardiac fats based on the other. This approach of correlation and prediction has not been previously addressed nor exploited in the literature. Furthermore, we also explore the feasibility of this approach, while comparing the performance of individual algorithms in the prediction task.

## 3. Materials and methods

We used the ground truth data available in [24,25] that was provided in our previous work [1]. The various images in the axial-plane were manually segmented by two experts. The epicardial and mediastinal fats in the images are represented as red and green pixels, respectively. The transition between the two (which also corresponds to the pericardium) is depicted in blue, as shown in the images of Fig. 2. Every non-black pixel in the image represents fat. In other words, black pixels represent everything that is not fat. More details on this fat range conversion can be found in [1,26].

The cardiac CT scan is composed of approximately 50 images. The amount of fat in these images varies depending on the slice number. In this work, we address the fact that the volumes of these two fats strongly correlate. Firstly, the images are converted to standard spacing in xyz-space so that the heart of each patient is approximately at the same scale. The term approximately refers to the fact that we use scanners from two manufacturers, which introduces variability.

As stated, blue pixels represent the pericardium, which is a very thin layer between the two fats. Since this information is already available in the ground truth, we decided to include it as a variable in the prediction. However, as the amount of blue pixels is very low, we would like to highlight some points about their treatment, as follows: (1) they can be ignored in the associated equations, if desired, (2) they can be considered either green or red pixels, (3) or even both red and green at the same time, as in our previous work [1]. Furthermore, (4) they could also be estimated by tracing the pericardium contour. The best approach depends on the task at hand and the requirements of the user.

It is important to emphasize that for predicting the epicardial fat (red pixels), the information we have *a priori* is depicted in Fig. 3-(a), where epicardial pixels are originally grey. Similarly, Fig. 3-(b) shows the *a priori* information when we wish to predict the amount of mediastinal (green) pixels.

Fig. 4 shows the input amount of grey pixels in these two prediction cases. Fig. 4-(a) corresponds to the scenario where we wish to predict the amount of epicardial (red) pixels. It can be seen in that image that epicardial pixels are originally grey, since this information is not available prior to processing. Fig. 4-(b), on the other hand, corresponds to the scenario where we wish to predict the amount of mediastinal (green)

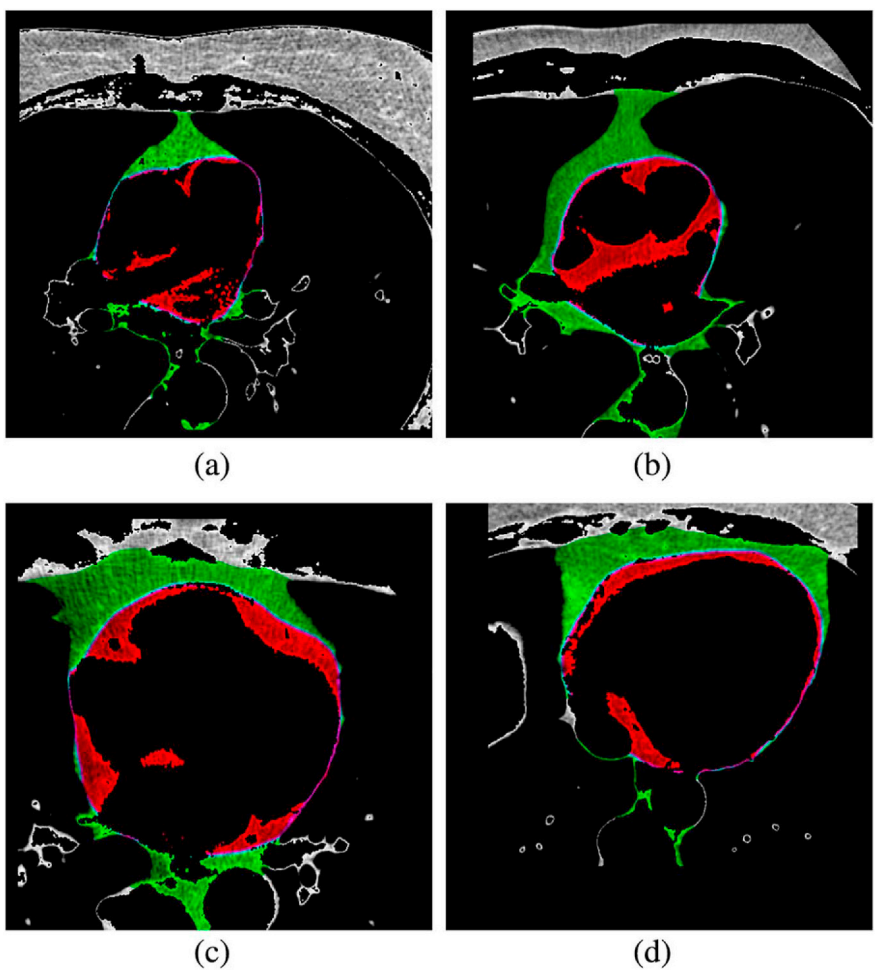

**Fig. 2.** Slices of individual patients containing only fat (red=epicardial, green=mediastinal, blue=pericardium and grey=other types of fat). (For interpretation of the references to color in this figure legend, the reader is referred to the web version of this article).

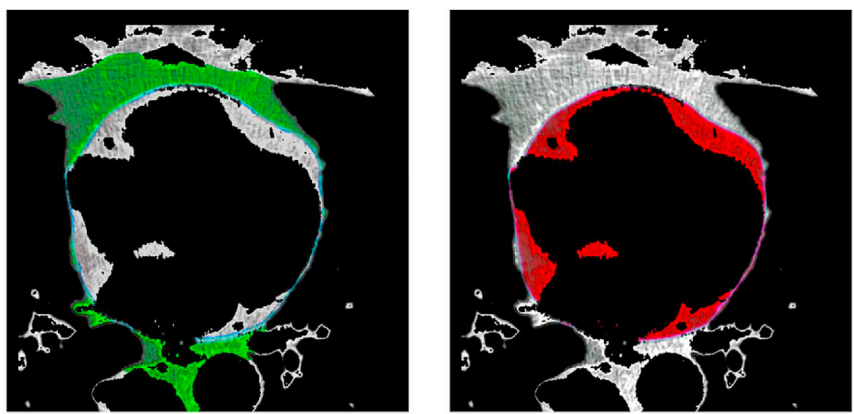

(a) Input images when predicting epicardial (red) pixels

(b) Input images when predicting mediastinal (green) pixels

**Fig. 3.** Example of input images for the prediction of epicardial and mediastinal fat. (a) How input images look like when predicting epicardial (red) pixels (b) How input images look like when predicting mediastinal (green) pixels. (For interpretation of the references to color in this figure legend, the reader is referred to the web version of this article).

pixels. Fig. 4-(c) and -(d) correspond to possible representations of the predictions.

The ground truth contains CT scans from 20 patients, which gives us a total of 878 images. For every image in the ground truth, we extract an instance of the dataset, which is composed of 7 features: (1) the amount of red pixels in the image, (2) the amount of green pixels in the image, (3) the amount of blue pixels in the image, (4) the amount of grey pixels in the image, (5) the amount of black pixels in the image (background), (6) the total amount of images of the associated scan and (7) the index of the slice to be processed (considering a scanning direction from head to feet).

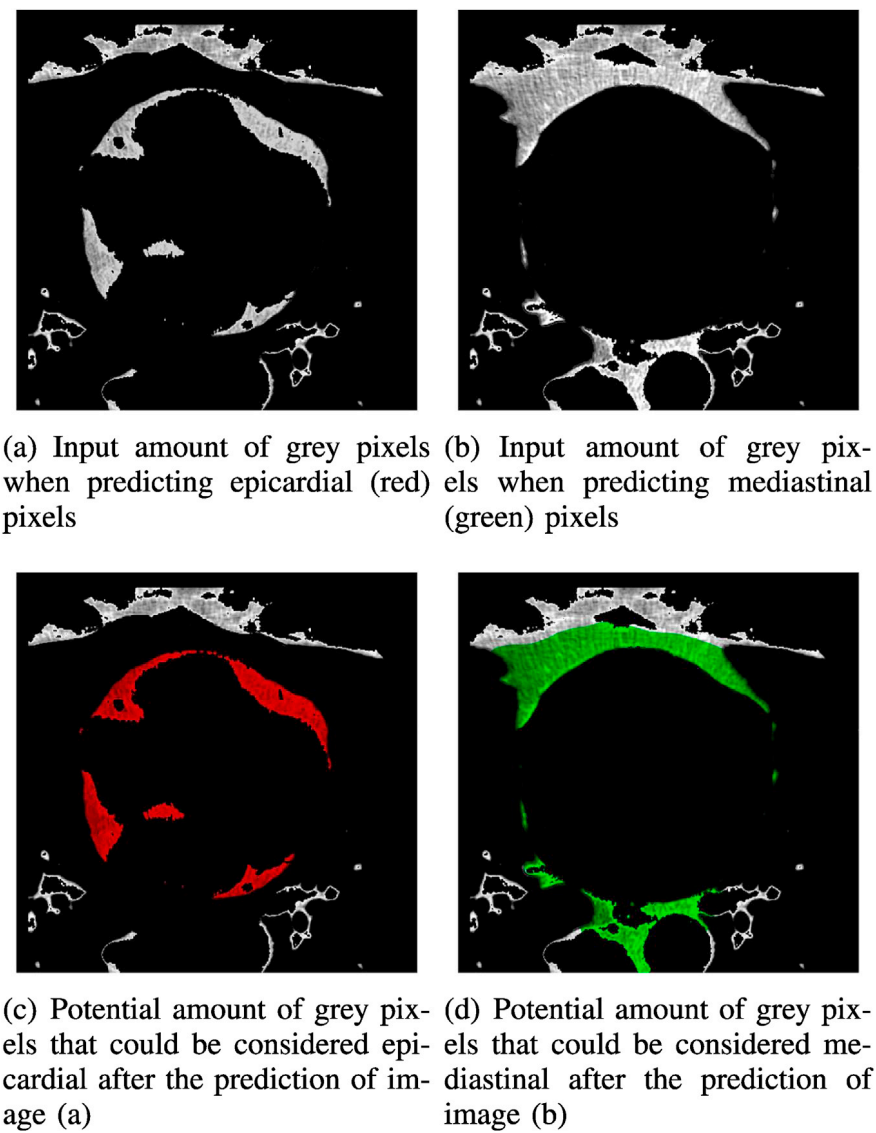

(a) Input amount of grey pixels when predicting epicardial (red) pixels

(b) Input amount of grey pixels when predicting mediastinal (green) pixels

(c) Potential amount of grey pixels that could be considered epicardial after the prediction of image (a)

(d) Potential amount of grey pixels that could be considered mediastinal after the prediction of image (b)

**Fig. 4.** Example of the amount of grey pixels when predicting the epicardial and mediastinal fat. (a) Input amount of grey pixels when predicting epicardial (red) pixels (b) Input amount of grey pixels when predicting mediastinal (green) pixels (c) Potential amount of grey pixels that could be considered epicardial after the prediction of image (a) (d) Potential amount of grey pixels that could be considered mediastinal after the prediction of image (b). (For interpretation of the references to color in this figure legend, the reader is referred to the web version of this article).

These 7 features are the attributes of the instances in our extracted dataset.

Although the extracted features may seem fairly simple in nature, they suffice for the purposes of prediction, as demonstrated in our experiments. Our approach considers the amount of fat in a specific slice so as to minimize errors that would appear if we attempted to predict the entire volume at once. Furthermore, it offers an additional advantage as it counts the amount of pixels corresponding to each fat at each slice, thus permitting the analysis of the correlation as processing evolves along the axial plane.

After generating the dataset composed of 878 instances (one for each image), we used the Weka framework [27] to perform the experiments. The class we wish to predict is continuous and, thus, the problem is posed as regression. We analyse the performance of the various algorithms based on the correlation coefficient, mean absolute error (MAE), root mean squared error (RMSE), relative absolute error (RAE) and root relative squared error (RRSE). The Pearson correlation coefficient $\rho$ is defined in Eq. (1):

$$\rho = \frac{\sum_{i=1}^{n}(a_i - \overline{a})(b_i - \overline{b})}{\sqrt{\sum_{i=1}^{n}(a_i - \overline{a})^2 \sum_{i=1}^{n}(b_i - \overline{b})^2}} \quad (1)$$

where $a_i$ stands for the predicted values and $b_i$ stands for the original values of the dataset, both with regards to instance $i$. $\overline{a}$ and $\overline{b}$ represent the mean of the predicted and original values, and $n$ is the total number of instances.

The mean absolute error (MAE) is given in Eq. (2) and measures the average error between the predicted and the original values:

$$MAE = \frac{1}{n}\sum_{i=1}^{n}|a_i - b_i| \quad (2)$$

The root mean squared error is a very similar metric to MAE and is shown in Eq. (3):

$$RMSE = \sqrt{\frac{1}{n}\sum_{i=1}^{n}(a_i - b_i)^2} \quad (3)$$

The relative absolute error (RAE) and root relative squared error (RRSE) are essentially the MAE and RMSE, but in relation to the mean. Their formulations are given by Eqs. (4) and (5), respectively:

$$RAE = \frac{\sum_{i=1}^{n} |a_i - b_i|}{\sum_{i=1}^{n} |\bar{b} - b_i|} \tag{4}$$

$$RRSE = \sqrt{\frac{\sum_{i=1}^{n} (a_i - b_i)^2}{\sum_{i=1}^{n} (\bar{b} - b_i)^2}} \tag{5}$$

### 3.1. Regression algorithms

Let the instances of our problem be described in a $\mathbb{R}^k$ space, where $k$ is the total number of features of each and every instance. The regression problem consists of finding the vector of weights $w$ that approximates the most the attribute values to the value we seek to predict (usually one of the attributes, say attribute $k$ represents the value we want to predict). The predicted or estimated value for a given instance $t^{(i)}$ is given by Eq. (6), where $t_j^{(i)}$ represents the attribute or dimension $j$ of instance $t^{(i)}$:

$$\sum_{j=1}^{k-1} w_j t_j^{(i)} = w_1 t_1^{(i)} + w_2 t_2^{(i)} + \cdots + w_{k-1} t_{k-1}^{(i)} \tag{6}$$

The regression goal is to minimize the total error among all the instances of the dataset. Thus, we wish to minimize the error $E$ on the training data, which is given by Eq. (7):

$$E = \sum_{i=1}^{n} \left| t_k^{(i)} - \sum_{j=1}^{k-1} w_j t_j^{(i)} \right| \tag{7}$$

There are many algorithms to determine the optimal set of weights, so as to minimize the error. However, the previous steps effectively summarize the methodology to be followed. A popular approach to solve regression problems are neural networks. These are function-based classifiers (similarly to linear regressors) and the computation of the weights can be performed using back propagation, for instance.

Whilst it may not be very intuitive, some decision tree-based algorithms can also perform regression really well (e.g., Random Forests algorithm [28]). They essentially discretize the data and treat the problem as a multi-label classification. Some lazy algorithms such as k-Nearest Neighbours (k-NN or IBk), which analyse the surrounding $k$ instances in order to classify an unlabeled instance, are also fairly good regression algorithms.

The main issue that raises from applying k-NN, neural networks or decision tree algorithms to a regression problem is that the predictive models produced by these algorithms are frequently rather complex for humans to understand. In this respect, linear regression algorithms are more appropriate as they are easier to interpret and to reproduce, since the resulting predictive model is a simple linear function.

## 4. Experimental results

At first, we analyse the performance of several algorithms for predicting the amount of mediastinal fat based on the amount of epicardial fat and the remaining features. That is, among the 7 previously described features, the amount of green pixels (mediastinal fat) is the class of the problem (Section 4.1). In the next group of experiments (Section 4.2), the amount of red pixels (epicardial fat) is the actual class of the problem.

One may argue that it is not necessary to perform these two types of experiments, since if we have a single function that relates both epicardial and mediastinal fats, it can be used to predict either one. Although this is true for linear regression algorithms (since the resulting model is a linear function), this is not generally true for other types of non-linear algorithms (e.g., decision tree, neural networks, etc.). Clearly, it is not straightforward to "invert" a predictive model that is intrinsically complex. In this research we perform these two groups of experiments separately.

### 4.1. Predicting the amount of mediastinal fat based on the amount of epicardial fat

The indices shown in Tables 1, 3, 5 and 6 were obtained using 10-fold cross validation. The parameters of each algorithm were experimentally adjusted (two to three combinations of parameters were tested for each algorithm) and the best results were selected. We also set an upper time limit of 10 min for running each of the algorithms (including the cross validation testing time).

It is possible to see in Table 1 that the Rotation Forest [29] with the Multilayer Perceptron [30] (MLP) Regressor as one of its parameters was the best performer in this experiment, achieving a correlation coefficient of 0.9876. The plot in Fig. 5 shows the actual amount of mediastinal fat (green pixels) on the $x$-axis versus the predicted or estimated amount of mediastinal fat on the $y$-axis, using the Rotation Forest+MLP Regressor.

Although the Linear Regression algorithm did not perform as well as other algorithms, this is probably the only algorithm among the ones in Table 1 that generates a model that can be easily understood. The correlation achieved with the Linear Regression algorithm is shown in Fig. 6, and can be compared to Fig. 5, which corresponds to the best result. The function generated by the Linear Regression on the full training dataset is shown in Eq. (8):

$$\begin{aligned} greenQnt = &-1.2295 \times redQnt - 7.4448 \times blueQnt \quad -0.9017 \\ &\times blackQnt - 72.8534 \times imageIndex \quad +233.0906 \\ &\times imagesQnt + 230102.0526 \end{aligned} \tag{8}$$

where $greenQnt$ represents the amount of mediastinal fat (green pixels) in the current slice ($imageIndex^{th}$ slice), $redQnt$ represents the amount of epicardial fat (red pixels) in the slice, $blueQnt$ represents the amount of blue pixels, $blackQnt$ the amount of black pixels, $imageIndex$ represents the number of the slice and $imagesQnt$ represents the total number of slices in the patient scan.

Once the amount of mediastinal fat pixels in each slice of a patient scan is estimated, it is trivial to obtain the total volume of this fat by accessing the xyz physical spacing information in the DICOM file and performing the conversion to a volume. The function in Eq. (8) must be computed for every slice of the scan. The only constant in this equation is *imagesQnt*, which does not change with regards to a single patient. The amount of grey pixels in the image was not relevant enough to be included in the prediction function, when using the linear regression algorithm.

**Table 1**
Comparison of individual regression algorithms for predicting the amount of mediastinal fat.

| Algorithm | $\rho$ | MAE | RMSE | RAE | RRSE |
|---|---|---|---|---|---|
| Rotation Forest+MLP Regressor | 0.9876 | 1331.8 | 1948.2 | 14.4 | 15.7 |
| RBF Regressor | 0.9866 | 1247.5 | 2027.1 | 13.7 | 16.1 |
| MLP Regressor | 0.9856 | 1449.1 | 2101.5 | 15.9 | 16.9 |
| SMO Regressor | 0.9852 | 1261.3 | 2130.4 | 13.8 | 17.1 |
| Rotation Forest+Random Forest | 0.9822 | 1663.3 | 2387.4 | 18.2 | 19.2 |
| Additive Regression+Random Forest | 0.9812 | 1596.9 | 2415.1 | 17.5 | 19.4 |
| k-NN/IBk | 0.9804 | 1604.5 | 2443.0 | 17.6 | 19.6 |
| Random Forest | 0.9763 | 1793.2 | 2700.3 | 19.7 | 21.7 |
| M5P | 0.9695 | 2356.7 | 3040.4 | 25.9 | 24.5 |
| Alternating Model Tree | 0.969 | 2286.0 | 3059.2 | 25.1 | 24.6 |
| M5 Rules | 0.9681 | 2375.0 | 3104.8 | 26.1 | 25.0 |
| Linear Regression | 0.9534 | 2875.3 | 3736.7 | 31.6 | 30.1 |
| Extra Tree | 0.9516 | 2487.8 | 3823.2 | 27.3 | 30.8 |
| LeastMedSq | 0.9493 | 2896.2 | 3944.1 | 31.8 | 31.7 |
| Elastic Net | 0.949 | 2977.8 | 3900.9 | 32.7 | 31.4 |
| REP Tree | 0.9431 | 2671.9 | 4139.3 | 29.3 | 33.3 |
| Random Tree | 0.9343 | 2747.7 | 4433.0 | 30.2 | 35.7 |

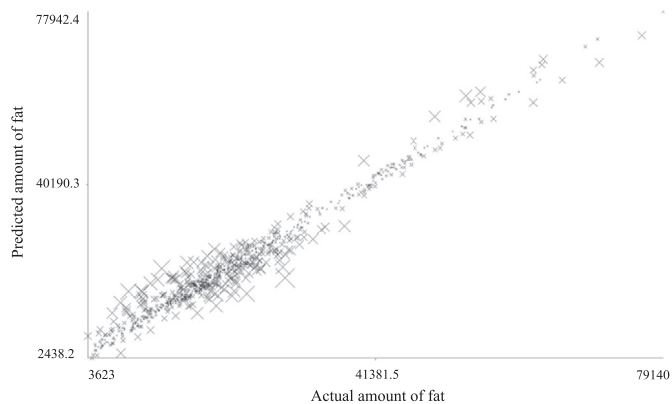

**Fig. 5.** Mediastinal fat correlation using Rotation Forest+MLP Regressor. Each point represents the amount of pixels in a single slice.

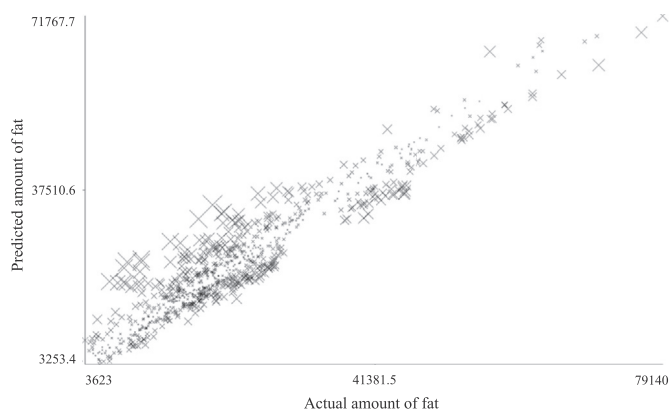

**Fig. 6.** Mediastinal fat correlation using Linear Regression algorithm. Each point represents the amount of pixels in a single slice.

Table 2 shows 10 randomly selected slices from the 20 patients. The first column corresponds to the actual amount of mediastinal fat in the slice (amount of pixels), the second column corresponds to the predicted amount using the Rotation Forest and MLP Regressor, while the third column shows the error between the actual and estimated amounts.

**Table 2**
Comparison of prediction results for mediastinal fat.

| Actual quantity | Predicted quantity | Error |
|---|---|---|
| 15125 | 15537.926 | 412.926 |
| 12660 | 13987.862 | 1327.862 |
| 11756 | 11935.518 | 179.518 |
| 10561 | 12157.081 | 1596.081 |
| 23418 | 24248.377 | 830.377 |
| 8673 | 10570.37 | 1897.37 |
| 11542 | 13753.043 | 2211.043 |
| 45177 | 44580.362 | −596.638 |
| 24319 | 24579.283 | 260.283 |
| 23270 | 19815.134 | −3454.866 |

## 4.2. Predicting the amount of epicardial fat based on the amount of mediastinal fat

Table 3 compares the performances of individual regression techniques on the prediction of the amount of epicardial fat. The performance of the best algorithms in Table 1 is similar to those in Table 3. However, the ranking of the algorithms changed substantially. Overall, the indexes are lower than in the previous table. This suggests that it is more difficult to predict the amount of epicardial fat given the mediastinal than the opposite.

It is worth emphasizing that the Linear Regression algorithm performed substantially worse when compared to the mediastinal fat prediction (Table 1). The resulting linear predictor trained on the entire dataset is shown in Eq. (9):

$$redQnt = -0.4608 \times greenQnt - 1.3373 \times blueQnt - 0.4736$$
$$\times blackQnt - 54.5244 \times imageIndex + 47.9363$$
$$\times imagesQnt + 123509.7603 \tag{9}$$

As previously mentioned, the regression algorithms did not perform as well in this class of the problem. Thus, we investigated whether inverting Eq. (8) could result in better performance. The Linear Regression algorithm in the mediastinal fat experiment achieved a correlation coefficient of 0.9534, which is fairly good in comparison to the correlation results in Table 3. Thus, if one wishes to predict the amount of epicardial fat given the mediastinal fat data, it could be better to consider the linear function of Eq. (8). Therefore, we experimented with the inversion of Eq. (8), which resulted in Eq. (10). However, when applying Eq. (10) to the problem, we obtain a correlation coefficient of 0.82369, where $MAE = 2319.5983$, $RMSE = 3013.6908$, $RAE = 55.95\%$ and $RRSE = 56.70\%$. These indices are slightly worse than the ones obtained with the Linear Regression algorithm, which are shown in Table 3.

$$redQnt = -0.8133 \times greenQnt - 6.0551 \times blueQnt - 0.7334$$
$$\times blackQnt - 59.2545 \times imageIndex + 189.5816$$
$$\times imagesQnt + 187150.9171 \tag{10}$$

Thus, we concluded that, based on the available evidence, it is more appropriate to use Eq. (9) rather than Eq. (10).

Furthermore, Figs. 7 and 8 show the correlations of the best performing algorithm in this experiment (Rotation Forest + MLP Regressor) and Linear Regression, respectively. The x and y axes represent the actual and predicted values of the amount of epicardial fat, respectively.

The predictive models generated in this experiment predict negative values in some slices as it can be seen in Figs. 7 and 8, which does not make sense in practice. This was not the case in the mediastinal fat experiment. Moreover, the spreading of the points in the plot is much wider in comparison to the previous experiment. The plots, however, have different scales, readily observable by the RAE and RRSE indices in Table 3.

In practice, the problem of negative values for the amount of epicardial fat could be resolved through normalization. Indeed, normalization could even further improve the accuracy of the obtained results. However, one of the intended purposes of this research is to propose a simple methodology for non-technical experts, where by directly inputting the information related to the amount of pixels for one fat, it is possible to obtain a fairly accurate prediction for the other type of fat, without any further need for processing. It may also be the case that the patient data could be outside of the normalization range. For instance, let us suppose that the maximum value for a certain attribute in the available training dataset is *m*. Thus, when the data is normalized, the maximum value for the particular attribute will be equal to 1. However, there may be instances where patients have attribute values higher than *m*, which was the available maximum on the training set. If this was to occur, regression models would not be able to provide sensible predictions for these instances. Thus, we decided to disregard the application of normalization as a rectification procedure.

**Table 3**
Comparison of individual regression algorithms for predicting the amount of epicardial fat.

| Algorithm | $\rho$ | MAE | RMSE | RAE | RRSE |
|---|---|---|---|---|---|
| Rotation Forest+MLP Regressor | 0.9683 | 671.0624 | 1095.029 | 19.6221 | 24.958 |
| RBF Regressor | 0.9592 | 736.551 | 1241.594 | 21.537 | 28.2985 |
| SMO Regression | 0.9569 | 783.9753 | 1271.063 | 22.9237 | 28.9702 |
| MLP Regressor | 0.9558 | 815.2913 | 1293.702 | 23.8394 | 29.4862 |
| Rotation Forest+Random Forest | 0.9497 | 964.0111 | 1434.355 | 28.188 | 32.6918 |
| Additive Regression+Random Forest | 0.9472 | 880.1396 | 1412.651 | 25.7355 | 32.1973 |
| k-NN/Ibk | 0.944 | 871.8146 | 1446.764 | 25.4921 | 32.9748 |
| Random Forest | 0.9389 | 997.4846 | 1538.362 | 29.1667 | 35.0625 |
| M5 Rules | 0.8783 | 1505.644 | 2092.582 | 44.0255 | 47.7005 |
| Alternating Model Tree | 0.8776 | 1492.431 | 2102.981 | 43.6391 | 47.9313 |
| M5P | 0.8702 | 1557.657 | 2160.526 | 45.5464 | 49.2429 |
| Extra Tree | 0.86 | 1414.435 | 2287.365 | 41.3585 | 52.1338 |
| Linear Regression | 0.8531 | 1724.72 | 2284.188 | 50.4313 | 52.064 |
| Random Tree | 0.8477 | 1432.02 | 2428.647 | 51.8727 | 55.3539 |
| Elastic Net | 0.8438 | 1750.837 | 2349.17 | 51.195 | 53.5425 |
| REP Tree | 0.8436 | 1566.34 | 2371.547 | 45.8003 | 54.0525 |
| LeastMedSq | 0.7929 | 1935.701 | 2748.017 | 56.6005 | 62.633 |

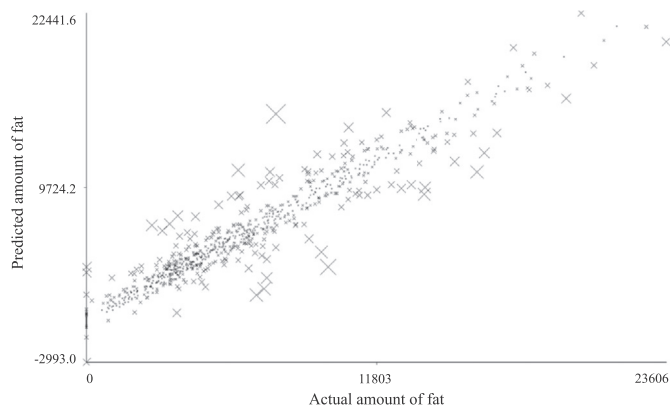

**Fig. 7.** Epicardial fat correlation using the Rotation Forest+MLP Regressor. Each point represents the amount of pixels in a single slice.

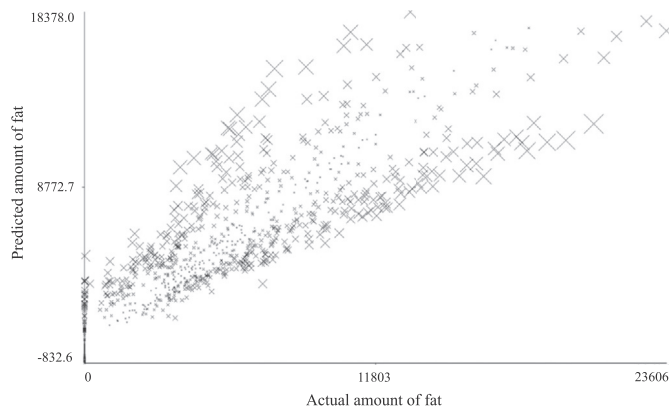

**Fig. 8.** Epicardial fat correlation using the Linear Regression algorithm. Each point represents the amount of pixels in a single slice.

Table 4 shows 10 randomly selected slices from the 20 patients. The first column corresponds to the actual amount of epicardial fat in the slice (amount of pixels), the second column corresponds to the predicted amount using the Rotation Forest and MLP Regressor, while the third column shows the error between the actual and estimated amounts.

**Table 4**
Comparison of prediction results for epicardial fat.

| Actual quantity | Predicted quantity | Error |
|---|---|---|
| 14506 | 13133.203 | −1372.797 |
| 7429 | 8579.046 | 1150.046 |
| 6804 | 5140.564 | −1663.436 |
| 7474 | 8158.426 | 684.426 |
| 3357 | 4088.433 | 731.433 |
| 10729 | 9287.318 | −1441.682 |
| 15068 | 15509.504 | 441.504 |
| 11724 | 10216.715 | −1507.285 |
| 6919 | 5005.346 | −1913.654 |
| 9110 | 7520.698 | −1589.302 |

### 4.3. Predicting both fats from unprocessed exams

For the sake of comparison, we perform a third and final experiment where the goal is to predict the epicardial and mediastinal fat volumes without preprocessing the images. That is, we investigate the prediction of the volume of the epicardial/mediastinal fat with no *a priori* information of the mediastinal/epicardial fat and the pericardium. Thus, the images of the patients would be entirely in greyscale, as shown in Fig. 9. Therefore, the pixels in these images are either grey, corresponding to fat, or black, depicting the background.

Table 5 compares the results obtained with each algorithm in this experiment. It is possible to conclude that, with regards to Table 3, the reported errors have almost doubled in most occasions. This confirms the validity of the hypothesis that it is more accurate to predict one type of fat based on the other, instead of attempting to predict it directly from unprocessed fat data. Despite the fact that the resulting models are not as accurate as in the case of the previous experiments, nevertheless, they provide acceptable results for the purposes of prediction. Indeed, this approach could be useful if the volumes of both fats are unknown and there is a requirement for rapid fat volume estimation.

Table 6 compares the results obtained when predicting the epicardial volume without any information of the mediastinal fat and pericardium. This experiment produced the worst results, the errors have nearly doubled, as shown in Table 5. In terms of medical relevance, accurate estimation of the epicardial fat is more significant than that of the mediastinal volume.

Figs. 10 and 11 show the correlation plots obtained for the best performing algorithm (Rotation Forest+MLP Regressor) for the mediastinal and epicardial fat, respectively. As expected, it is evident that errors are much higher than the ones in Sections 4.1 and 4.2. More specifically, the plot of Fig. 10 is more spread than that of Fig. 5 and the plot of Fig. 11 is more spread than that of Fig. 7.

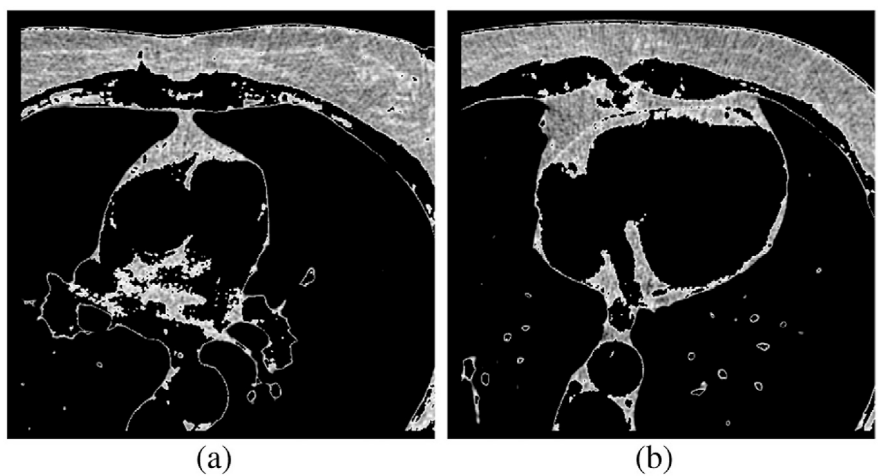

**Fig. 9.** Examples of unprocessed slices of individual patient CT examinations.

**Table 5**
Comparison of individual regression algorithms for prediction of the amount of mediastinal fat on unprocessed images.

| Algorithm | $\rho$ | MAE | RMSE | RAE | RRSE |
|---|---|---|---|---|---|
| Rotation Forest+MLP Regressor | 0.9549 | 2464.511 | 3681.298 | 27.1063 | 29.6706 |
| k-NN/Ibk | 0.9544 | 2102.944 | 3701.523 | 23.1295 | 29.8336 |
| MLP Regressor | 0.9531 | 2502.514 | 3749.523 | 27.5243 | 30.2205 |
| RBF Regressor | 0.946 | 2366.389 | 4030.551 | 26.0271 | 32.4856 |
| Random Forest | 0.9456 | 2624.493 | 4031.234 | 28.8659 | 32.4911 |
| Additive Regression+Random Forest | 0.9448 | 2532.347 | 4063.312 | 27.8524 | 32.7496 |
| SMO Regression | 0.942 | 2815.128 | 4176.027 | 30.9626 | 33.6581 |
| RotationForest+Random Forest | 0.942 | 2804.584 | 4180.239 | 30.8466 | 33.692 |
| Alternating Model Tree | 0.9202 | 3760.431 | 4871.069 | 41.3596 | 39.26 |
| REP Tree | 0.9081 | 3521.787 | 5224.702 | 38.7349 | 42.1102 |
| M5P | 0.9076 | 3875.261 | 5231.124 | 42.6226 | 42.162 |
| M5 Rules | 0.903 | 3940.991 | 5330.788 | 43.3456 | 42.9652 |
| Random Tree | 0.8983 | 3364.667 | 5644.721 | 37.0068 | 45.4955 |
| Extra Tree | 0.8975 | 3385.886 | 5569.347 | 37.2401 | 44.888 |
| Linear Regression | 0.8144 | 5525.071 | 7189.585 | 60.7683 | 57.9468 |
| LeastMedSq | 0.8121 | 5461.548 | 7234.044 | 60.0696 | 58.3052 |
| Elastic Net | 0.7945 | 5778.268 | 7522.813 | 63.5531 | 60.6362 |

**Table 6**
Comparison of individual regression algorithms for prediction of the amount of epicardial fat on unprocessed images.

| Algorithm | $\rho$ | MAE | RMSE | RAE | RRSE |
|---|---|---|---|---|---|
| RotationForest+ MLPRegressor | 0.9283 | 1112.602 | 1630.195 | 32.5338 | 37.1155 |
| MLP Regressor | 0.921 | 1170.427 | 1709.535 | 34.2236 | 38.9639 |
| k-NN/Ibk | 0.9186 | 1026.236 | 1733.968 | 30.0075 | 39.5207 |
| RBF Regressor | 0.9129 | 1111.812 | 1796.067 | 32.5097 | 40.9361 |
| Additive Reg.+Random Forest | 0.9038 | 1210.713 | 1874.056 | 35.4016 | 42.7136 |
| Random Forest | 0.8999 | 1280.327 | 1922.779 | 37.4371 | 43.8241 |
| RotationForest+Random Forest | 0.8915 | 1407.415 | 2036.57 | 41.1532 | 46.4177 |
| SMO Regression | 0.8546 | 1554.29 | 2312.397 | 45.4479 | 52.7044 |
| Extra Tree | 0.8366 | 1556.863 | 2459.925 | 45.5231 | 56.0668 |
| Alternating Model Tree | 0.8126 | 1957.636 | 2553.028 | 57.2419 | 58.1888 |
| M5P | 0.7952 | 1972.226 | 2680.816 | 57.6688 | 61.1014 |
| Random Tree | 0.7884 | 1724.339 | 2801.167 | 50.4202 | 63.8444 |
| REP Tree | 0.7651 | 1927.994 | 2843.145 | 56.3751 | 64.8012 |
| M5Rules | 0.7494 | 2087.658 | 2901.227 | 61.0437 | 66.125 |
| Linear Regression | 0.4907 | 2815.204 | 3814.83 | 82.3174 | 86.9479 |
| LeastMedSq | 0.4455 | 2816.671 | 3948.312 | 82.3603 | 89.9903 |
| Elastic Net | 0.4218 | 2878.6 | 3969.828 | 84.1711 | 90.4807 |

## 5. Conclusions

In this research, we proposed the use of regression algorithms for predicting epicardial and/or mediastinal fats in relation to one another and to the image data. Two numerical values representing the volumes of the epicardial and mediastinal fats are the main interest in most medical analyses. Therefore, visual segmentation of the images can be bypassed if there exists a reliable way of estimating the required fat volume, such as the approach proposed in this contribution.

In a previous work [1], full segmentation of epicardial and mediastinal fat volumes took 1.8 h. Our experiments demonstrated that there is no strict need to process and segment both fats, but rather to only process one instead. Indeed, this implies some additional errors. However, the obtained results demonstrate a high correlation between the segmented fat volumes and the estimated ones. The key benefit of this approach is that processing time is reduced by half, to approximately 0.9 h, using this methodology.

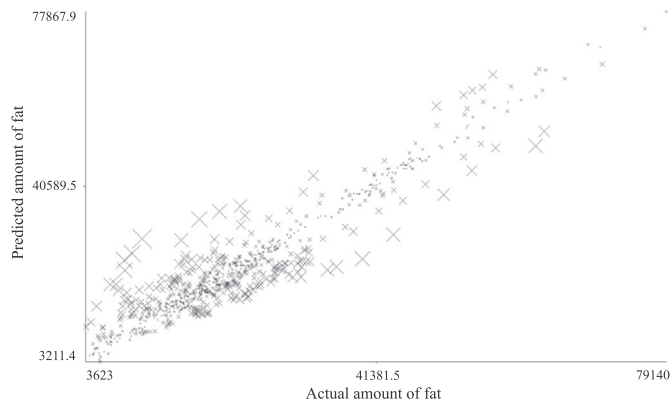

**Fig. 10.** Mediastinal fat correlation using the Rotation Forest+MLP Regressor in the unprocessed images experiment. Each point represents the amount of pixels in a single slice.

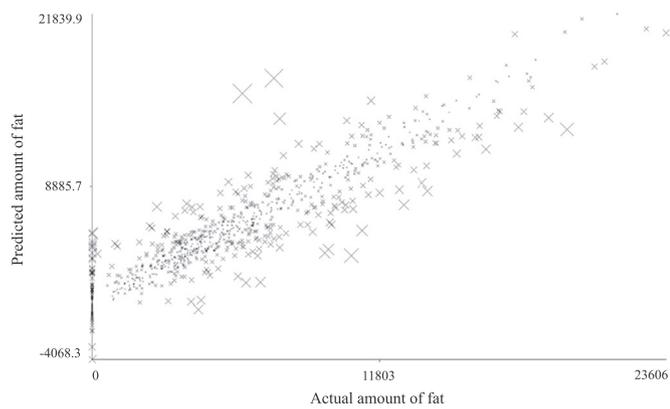

**Fig. 11.** Epicardial fat correlation using the Rotation Forest+MLP Regressor in the unprocessed images experiment. Each point represents the amount of pixels in a single slice.

With regards to the reported results, it is preferable to predict the mediastinal fat volume based on the epicardial than the opposite. The correlation coefficient obtained by the Rotation Forest algorithm using the MLP Regressor as parameter for predicting the mediastinal fat was 0.9876, with a relative absolute error of 14.4% and root relative squared error of 15.7%.

The best correlation coefficient obtained for predicting the epicardial fat based on the mediastinal was 0.9683 with relative absolute error of 19.6% and relative squared error of 24.9%. The maximum correlation coefficient achieved when predicting the mediastinal fat with no *a priori* information of the epicardial and pericardium was 0.9549 with a RAE of 27.1% and a RRSE of 29.6%, which is almost double the average error obtained when information of the epicardial fat is considered or known *a priori*. In the case of the epicardial volume prediction, the correlation coefficient was 0.9283 with a RAE of 32.5% and a RRSE of 37.1%. Once again, approximately twice the average error obtained when the mediastinal volume information is used.

The produced errors arise due to various factors. Examples of these are the lack of a very accurate registration over the entire patient dataset, the use of different CT scanners, exceptions to the norm, and indeed the prediction power of the employed regression algorithms. Registration was performed in the ground truth images, but this was an automated procedure, proposed in our previous work [1] and inadvertently introduces errors. A manual, more accurate registration would reduce the errors produced on the prediction. Furthermore, if the CT scans are calibrated to each other, acquired with the same parameters and the same device as well as registered with an accurate registration, then the errors could be substantially reduced.

We conclude that it is possible to predict fairly confidently one type of fat based on the information of the other. Overall, the obtained results are encouraging. Of particular note is that the utilized ground truth contains CT images from two manufacturers (Siemens and Philips), which implies that the proposed methodology (e.g., associated regression models) may be applicable irrespectively of the type of the CT scanners. Further research will concentrate on evaluating the application of the generated linear functions and predictive models to CT examinations obtained using scanners of other manufacturers to verify their generalization properties.

## Acknowledgments

The authors would like to thank CAPES-Brazil for providing scholarships and CNPQ-Brazil, grant number: 303240/20015-6.